\documentclass[aoas,MSNbibl,nameyear,dvips]{arximspdf}
\usepackage{dcolumn}
\usepackage[noend]{algorithmic}
\usepackage{algorithm}
\usepackage{graphicx}
\usepackage{breakurl}

\doi{10.1214/10-AOAS392}
\volume{5}
\issue{1}
\pubyear{2011}
\firstpage{254}
\lastpage{282}

\makeatletter
\newcommand{\comment}[1]{}
\newcommand{\VAR}[1]{\mbox{\textit{#1}}}
\newcommand{\FN}[1]{\mbox{\textsc{#1}}}
\newcommand{\NULL}{\textsc{Null}}
\newcommand{\Comment}[1]{$//$ \textit{#1}}

\renewcommand{\emptyset}{\varnothing}
\newcolumntype{d}[1]{D{.}{.}{#1}}
\newcommand{\eqref}[1]{(\ref{#1})}
\newcommand{\Ind}[1]{\mathbf{1}_{\{#1\}}}
\newcommand{\Exp}[1]{\operatorname{Exp}(#1)}
\newcommand{\old}{\operatorname{old}}
\newcommand{\new}{\operatorname{new}}
\newcommand{\NQ}[1]{N(u_{{#1}})}
\newcommand{\IQ}[1]{Q_{{#1}}}
\newcommand{\ts}{\Delta t}
\newcommand{\bs}{\mathbf{s}}
\newcommand{\bw}{\mathbf{w}}
\newcommand{\ba}{\mathbf{a}}
\newcommand{\bd}{\mathbf{d}}
\newcommand{\pdev}[2]{\frac{\partial #1}{\partial #2}}
\newcommand{\swname}[1]{\texttt{#1}}
\newcommand{\bq}{\mathbf{q}}
\newcommand{\calN}{\mathcal{N}}
\newcommand{\calT}{\mathcal{T}}

\newproclaim{remark}{Remark}
\makeatother

\begin{document}
\begin{frontmatter}

\title{Bayesian inference for queueing networks and modeling of
internet services}
\runtitle{Bayesian inference for queueing networks}

\begin{aug}
\author[A]{\fnms{Charles} \snm{Sutton}\corref{}\ead[label=e1]{csutton@inf.ed.ac.uk}}
\and
\author[B]{\fnms{Michael I.} \snm{Jordan}\ead[label=e2]{jordan@stat.berkeley.edu}}

\runauthor{C. Sutton and M. I. Jordan}

\affiliation{University of Edinburgh and University of California}

\address[A]{School of Informatics \\
University of Edinburgh \\
Edinburgh \\
EH8 9AB\\
UK \\
\printead{e1}} 

\address[B]{Department of Statistics \\
and Department of EECS\\
University of California \\
Berkeley, California 94720\\
USA \\
\printead{e2}}
\end{aug}

\received{\smonth{1} \syear{2010}}
\revised{\smonth{7} \syear{2010}}

%
\begin{abstract}
Modern Internet services, such as those at Google, Yahoo!, and Amazon,
handle billions of requests per day on clusters of thousands of
computers. Because these services operate under strict performance
requirements, a statistical understanding of their performance is of
great practical interest. Such services are modeled by networks of
queues, where each queue models one of the computers in the system. A
key challenge is that the data are incomplete, because recording
detailed information about every request to a heavily used system can
require unacceptable overhead. In this
paper we develop a Bayesian perspective on queueing models in which the
arrival and departure times that are not observed are treated as latent
variables. Underlying this viewpoint is the observation that a queueing
model defines a deterministic transformation between the data and a set
of independent variables called the service times. With this viewpoint
in hand, we sample from the posterior distribution over missing data
and model parameters using Markov chain Monte Carlo. We evaluate our
framework on data from a benchmark Web application. We also present a
simple technique for selection among nested queueing
models. We are unaware of any previous work that considers inference
in networks of queues in the presence of missing data.
\end{abstract}

%

\end{frontmatter}

\section{Introduction}

Modern Internet services, such as those at Google, Yahoo!, and Amazon,
serve large numbers of users simultaneously;
for example, both eBay and Facebook claim over 300 million users
worldwide.\footnote{eBay: Franco Travostino and Randy Shoup, Invited
talk at workshop on Large Scale Distributed Systems and Middleware
(LADIS) 2009, available at  \url
{http://www.facebook.com/press/info.php?statistics}. Retrieved 3 November 2009.}
To handle these demands, large-scale Web applications are run on
clusters of thousands of individual networked machines, allowing large
numbers of requests to be served by processing different requests in
parallel on different machines. Each individual machine also processes
multiple requests simultaneously, and a typical request involves
computation on many machines. Web services also operate under strict
performance requirements. It is extremely important to minimize a
site's \textit{response time}---that is, the amount of time required
for a Web page to be returned in response to the user's
request---because even small delays, such as 100 ms, are sufficient to
cause a measurable decrease in business.\footnote{For measurements of
this phenomenon at Google and Microsoft, see \url
{http://velocityconference.blip.tv/file/2279751/}. See also \url
{http://perspectives.mvdirona.com/2009/10/31/TheCostOfLatency.aspx}.}

Developers of Web applications are concerned with two types of
statistical questions about performance. The first involve prediction
of the response time of the application under new conditions, for
example, if ten more Web servers were to be added, or if the number of
users were to double. These extrapolation-type questions are crucial
for configuring systems and for attempting to assess the robustness of
a system to a spike in load. The second type of statistical question
involves diagnosing the cause of observed poor performance in the
system. For example, a Web service could run slowly because one
component of the system, such as a database, is overloaded, meaning
that it is receiving more requests than it can handle quickly. Or, an
alternative explanation is that the database is not overloaded, but is
too slow even at low request rates, for example, because it is
configured incorrectly. It is important to distinguish these two
potential diagnoses of a performance problem, because their remedies
are different.

Both hypothetical and post hoc questions can be answered using a
\textit{performance model}, which is essentially a regression of
system performance, such as the response time, onto the workload and
other covariates.
Perhaps the most popular models are networks of queues [e.g., \citet
{kleinrock}]. For example, in a Web service, it is natural to model
each server by a single queue, and connect the queues using our
knowledge of how requests are processed by the system.
Queueing networks allow analysts to incorporate two important forms of
qualitative prior knowledge: first, the structure of the queueing
network can be used to capture known connectivity, and second, the
queueing model inherently incorporates the assumption that the response
time explodes when the workload approaches the system's maximum capacity.
This allows the model to answer hypothetical extrapolation-type
questions in a way that simple regression models do not.

Two inferential tasks will be of primary concern in this work.
The first is inference about the parameters of the queueing network.
This allows answering the hypothetical extrapolation-type questions by
simulating from the network. For example, if we wish to know how well
the system will perform if the request rate doubles, we can simply
simulate from the queueing network with the inferred parameters but
doubling the arrival rate.
The second task is to infer for each request to the system how much
time was spent in the queue---this is called the
\textit{waiting time}---and how much time was spent in processing
after the request reached the head of the queue---this is called the
\textit{service time}. This allows us to infer if any system
components are overloaded, because requests to those components will
have large waiting times.

However, the inferential setup is complicated by the fact that Web
services operate under strict performance requirements, so that data
must be collected in a manner that requires minimal overhead.
Observation schemes whose overhead is trivial for a small number of
requests can cause unacceptable delay in a site that receives millions
of requests per day.
For this reason, {incomplete} observation schemes are common; for
example, the data set might contain the total number of requests that
arrive per second, and the response time for 10\% of the requests. The
goal behind such a scheme is to minimize the amount of computation and
storage required to collect
the data, at the expense of increasing the complexity of the analysis.

In this paper we introduce a novel inferential framework for networks
of queues with incomplete data. The main idea is to view a network of
queues as a direct model of the arrival and departure times of each
request, treating the arrival and departure times that have not been
measured as missing data.
Underlying this viewpoint is the observation that a queueing model
defines a deterministic transformation from a set of independent
variables, namely, the service times, to the arrival and departure
times, which can be observed. This deterministic transformation is
described in detail in Section~\ref{sec:modeling}. This perspective is
general enough to handle fairly advanced types of queueing models,
including general service distributions, multiprocessor queues and
processor-sharing queues [\citet{kleinrock}].
With this perspective in hand, the unmeasured arrival and departures
can be approximately sampled from their posterior distribution using
Markov chain Monte Carlo (MCMC). Once we can resample the arrival and
departure times, it is straightforward to estimate the parameters of
the network, either in a Bayesian framework or in a maximum likelihood
framework using Monte Carlo EM.

The design of posterior sampling algorithms presents technical
challenges that are specific to queueing models, mainly because the
missing arrival and departure times have complex deterministic
dependencies, for example, that each arrival time must be less than its
associated departure time, and that a departure time from one queue in
the network will necessarily equal an arrival time at some other queue
in the network.
We are unaware of previous work that considers missing data in networks
of queues.


\section{Modeling}
\label{sec:modeling}

\begin{figure}

\includegraphics{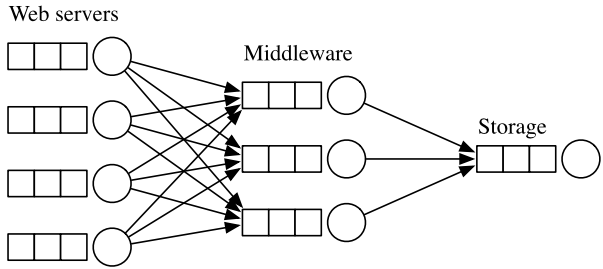}

\caption{A queueing network model of a three-tier Web service. The
circles indicate servers, and the boxes indicate queues.}
\label{fig:3tier}
\end{figure}

Many computer systems are naturally modeled as networks of queues.
Much work has been concerned with \textit{distributed systems} in
which the computation required by a single request is shared over a
large number of individual machines.
For example, Web services in particular are often designed in a
``three tier'' architecture (Figure~\ref{fig:3tier}), in which the
first tier is a presentation layer that generates the Web page
containing the response, the second tier performs application-specific logic,
and the third tier handles long-term storage, often using a database.
In order to handle the high request rates that are typical of a Web
service, each tier actually consists of multiple individual machines
that are equivalent in function. When a request for a Web page arrives
over the Internet, it is randomly assigned to one of the servers in the
first tier, which makes requests to the second tier as necessary to
generate the response. In turn, the second tier makes requests to the
third tier when data is required from long-term storage, for example,
data about a user's friends in a social networking site, or data about
Web pages that have been downloaded by a search engine.

%

It is natural to model a distributed system by a network of queues, in
which one queue models each individual machine in the system. The
queues are connected to reflect our knowledge of the system structure.
For example, in a Web service, we might model the processing of a
request as follows: Each request is randomly assigned to one of the
queues in the first tier, waits if necessary, is served, repeats this
process on the second and third tiers, and finally a response is
returned to the user.
(In reality, the first tier may call the second tier multiple times to
serve a single request, but we ignore this issue for modeling simplicity.)

Thus, each external request to the system might involve processing at
many individual queues in the network. To keep the terminology clear,
we will say that a \textit{job} is a request to one of the individual queues,
and a \textit{task} is the series of jobs that are caused by a single
external request to the system.
For example, consider a Web service that is modeled by the queueing
network in Figure~\ref{fig:3tier}. A task represents the entire
process of the system serving an external request that arrives over the
Web. A typical task in this model would comprise three jobs, one for
each tier of the system.

In order to define a probabilistic model over the arrival and departure
times of each job,
we need to model both (a) which queues are selected to process each job
in a task and (b)
the processing that occurs at each individual queue. For (a), we model
the sequence
of queues traversed by a task as a first-order Markov chain. We call
this sequence of
queues the \textit{path} of that task.
A task completes when the Markov chain reaches a designated final
state, so that
the number of jobs in each task is potentially random.

Second, to model the processing at individual queues, we consider
several different possibilities.
In the simplest model, each individual machine can process one job at a
time, in the order that they arrive;
this is called a \textit{single-processor first-come first-served
(FCFS)} queue.
In this model, jobs arrive at the system according to some point
process, such as a Poisson process. The \textit{interarrival times}
$\delta_e$ for each job $e$ are assumed to be drawn independently from
some density $g$. The arrival times themselves are denoted $a_e$. Once
a job arrives, it waits in the queue until all previous jobs have
departed. The amount of time spent in the queue is called the \textit
{waiting time}. Finally, once the job arrives at the head of the queue,
it spends some amount of time in processing, called the \textit
{service time} $s_e$, which we assume to be drawn from some density
$f$. The interarrival times and service times for all jobs are mutually
independent. Once the service time has elapsed, the job leaves the
system, and the next job can enter service. The departure time of job
$e$ is denoted $d_e$, and
the time that $e$ enters service is called the \textit{commencement
time} $u_e$. Finally, we use $\ba$ to represent the vector of arrival
times for all jobs, and, similarly, $\bd$ represents the vector of
departure times.

There is a more general way to view this model, which will be useful in
treating the more complex queueing disciplines considered later in this
section. In this view, we imagine that all of the interarrival times
$\delta_e$ and service times $s_e$ are drawn independently at the
beginning of time. Then the arrival and departure times are computed
from these variables via a deterministic transformation, which is given
by solving the system of equations
%
%
\begin{eqnarray}
a_e &=& \delta_e + a_{e-1}, \nonumber
\\[-8pt]
\\[-8pt]
d_e &=& s_e + \max[ a_e, d_{e-1} ].\nonumber
\end{eqnarray}
This transformation is one-to-one, so that observing all of the arrival
and departure times is equivalent to observing the i.i.d. service and
interarrival times.
In the remainder of this paper, the queueing regimes that we consider
are more complex, but still they can all be viewed in this general
framework, with different types of queues using different transformations.

The \textit{response time} $r_e$ of a job $e$ is simply the total
amount of time that the job requires to be processed, including both
waiting and service, that is, $r_e = d_e - a_e$. The \textit{waiting
time} $w_e$ of a job is the amount of time that the job spends in the
queue, that is, the response time minus the service time, so that $r_e
= w_e + s_e$. In this way, a queueing model can be interpreted as
decomposing the response time of a job into two components: the waiting
time, which models the effect of workload, and the service time, which
is independent of workload, and models the amount of processing which
is intrinsically required to service the job.

In the remainder of this section we describe three more sophisticated
queueing regimes from this perspective: multiprocessor first-come
first-served (FCFS) queues (Section~\ref{sec:fcfs}), queues which
employ random selection for service (RSS) (Section~\ref{sec:rss}), and
processor sharing (PS) queues (Section~\ref{sec:ps}). In all of those
sections we describe single-queue models.
Finally, in Section~\ref{sec:network} we place these single-queue
models within the context of a general framework for queueing networks.

\subsection{Multiprocessor FCFS queues}
\label{sec:fcfs}

In an FCFS queue, requests are removed from the queue in a first-come
first-served
(FCFS) manner. The queue is allowed to process $K$ requests
simultaneously, so no
requests need to wait in queue until $K+1$ requests arrive. This is
called a $K$-processor FCFS queue.

As before, the interarrival times $\delta_e$ are distributed
independently according to some density $g$, and
the resulting arrival times are defined as $a_e = a_{e-1} + \delta_e$
for all $e$.
The departure times are more complex.
First, the service times $s_{e} \sim f$ are distributed independently
of each other and the interarrival times.
Then, to transform the service times into departure times, observe that
a job enters service when at least one of the $K$ processors is free,
that is, when all but $K-1$ of the previous jobs have departed.
So we introduce auxiliary variables
$p_{e}$ to indicate which of the $K$ servers has been assigned to job
$e$, the time $b_{ek}$ to indicate the first time after job $e$ arrives
that the server $k$ would be clear, and $c_{e}$ to indicate the first
time after $e$ arrives that any of the $K$ servers are clear. Then the
departure times $d_{e}$ can be computed by solving the system of equations
%
%
\begin{eqnarray}
\label{eq:ggkfcfs:eqns}
b_{ek} &=& \max\{d_{e'} \mid a_{e'} < a_{e} \mbox{ and } p_{e'} = k
\}, \qquad
c_{e}  =  \min_{k \in[0,K)} b_{ek},\nonumber
\\[-8pt]
\\[-8pt]
p_{e} &=& \arg\min_{k \in[0,K)} b_{ek}, \qquad
u_{e}  = \max[ a_{e}, c_{e} ] ,  \qquad d_{e} = s_{e} + u_{e}.\nonumber
\end{eqnarray}
To obtain the joint density over arrival and departure times, we
require the Jacobian of the transformation $(\ba, \bd) \mapsto(\bs,
\delta)$ that maps the vector of arrival and departure times
to the i.i.d. interarrival and service times.
Fortunately, the Jacobian matrix $J$ of this map is triangular,
because any $a_{e}$ depends only on $\delta_1, \delta_2, \ldots,
\delta_e$, and any $d_{e}$ depends on $\delta_1, \delta_2, \ldots,
\delta_N$ and $s_1, s_2, \ldots, s_{e}$. So
\[
|\det J(\ba, \bd)| = \prod_{e=1}^{N} \biggl| \pdev{\delta
_{e}}{a_{e}} \biggr| \cdot\biggl| \pdev{s_{e}}{d_{e}} \biggr| = 1,
\]
where the jobs are indexed by $e \in[1,N]$.

The joint density over arrival and departure times is therefore
%
%
\begin{equation}
p(\ba, \bd) = \prod_{e=1}^N g (a_{e} - a_{e-1}) f (d_{e} -
u_{e})\label{eq:fcfs:lik}.
\end{equation}

\subsection{RSS queues}
\label{sec:rss}

In an RSS queue, when the processor finishes a job, the next job to be
processed is chosen randomly from all jobs currently in queue. (RSS
stands for Random Selection for Service.)
As before, interarrival and service times are generated from $f$ and
$g$ independently. To compute the arrival and departure times, define
$\gamma(e)$ as the predecessor of job $e$ in the departure order of
the queue and $\IQ{e}$ as the set of jobs in queue when $e$ departs.
Both these variables and the departure times can be computed from the
interarrival and service times by the system of equations
%
%
\begin{eqnarray}
\label{eq:gg1rss:eqns}
u_{e} &=& \max\bigl[ a_{e}, d_{\gamma(e)} \bigr], \qquad
\IQ{e}  = \{ e' \mid a_{e'} < d_{e} \mbox{ and } d_{e} < d_{e'} \},
\nonumber
\\[-8pt]
\\[-8pt]
d_{e} &=& s_{e} + u_{e}, \qquad
\gamma^{-1}(e)  =
\cases{\displaystyle
\operatorname{Random}(\IQ{e}), & \quad   if $\IQ{e} \neq\emptyset$
\cr\displaystyle
\arg\min_{\{ e' | d_{e} < a_{e'} \}} a_{e'}, & \quad   otherwise,
}\nonumber
\end{eqnarray}
where Random($S$) indicates an element of the set $S$, chosen uniformly
at random.
Notice that $\gamma(e)$ is always the job immediately preceding $e$ in
the departure order of the queue.

The likelihood for this model contains two types of factors: one that
arises from the service density, and one that arises from the random
selection of jobs from the queue. For the latter purpose, let $N(t)$ be
the number of jobs in the system at time~$t$, so that $\NQ{e}$ is the
number of jobs in queue when $e$ enters service, that is,
%
%
\begin{equation}
\label{eq:NQ}
\NQ{e} = 1 + \#\{ e' \mid a_{e'} < a_{e} \mbox{ and } u_{e} < d_{e'}
\}.
\end{equation}
Then the joint density over arrivals and departures is
%
%
\begin{equation}
\label{eq:rss:lik}
p(\ba, \bd) = \prod_{e=1}^N \NQ{e}^{-1} g(a_{e} - a_{e-1}) f(d_{e}
- u_{e}), 
\end{equation}
where, using similar reasoning to the FCFS case, it can be shown that
the Jacobian of the map $(\ba, \bd) \mapsto(\delta, \bs)$ is 1.

\subsection{Processor sharing queues}
\label{sec:ps}

A processor sharing (PS) queue [\citet{kleinrock}] is designed to model
computer systems that handle multiple jobs simultaneously on a single
processor via time sharing. One way to understand this queue is to
imagine the system as if it were in discrete time, with each time slice
consuming some time $\ts> 0$. When a job $e$ arrives at the queue, it
samples a total service time $s_{e} \in\Re$. Then, at each time slice
$t$, all of the $N(t)$ jobs remaining in the system have their service
times reduced by $\ts/ N(t)$. Once the remaining service time of a job
drops below zero, it leaves the queue.
The PS queue arises in the limit as $\ts\rightarrow0$. Intuitively,
each job in the system at any time $t$ instantaneously receives
$1/N(t)$ of the system's processing power.

Precisely, the PS queue defines a distribution over arrival and
departure times as follows.
First, the interarrival times $\delta_{e}$ are distributed
independently according to $g$, and the service times $s_e$
independently according to $f$.
Then, the arrival times are computed as $a_{e} = a_{e-1} + \delta_{e}$.
Finally, the departure times are defined as the solution to the system
of equations
%
%
\begin{eqnarray}
\label{eq:gg1ps:eqns}
N(t) &=& \sum_{e=1}^{N} \Ind{a_{e} < t}\Ind{t < d_{e}},\nonumber
\\[-8pt]
\\[-8pt]
s_{e} &=& \int_{a_{e}}^{d_{e}} \frac{1}{N(t)} \, dt.\nonumber
\end{eqnarray}
These equations can be solved iteratively by alternately holding the
function $N(t)$ fixed and solving the second equation for $d_{e}$, and
then holding $d_{e}$ fixed and solving the first equation for $N(t)$.
This procedure converges because $N(t)$ and all $d_{e}$ can only
increase at each iteration, and both are bounded from above.

The joint density in this case is complicated by a Jacobian term,
which, unlike in FCFS and RSS queues, does not vanish.
To compute the Jacobian, observe that $1/N(t)$ is a step function with
knots whenever a job arrives or departs. For a job~$e$, denote the
knots that occur in $[a_e, d_e]$ as $a_{e} = x_1 < x_2 < \cdots< x_M =
d_e$. So \eqref{eq:gg1ps:eqns} can be rewritten as
\[
s_{e} = \sum_{m=2}^{M} \frac{x_{m} - x_{m-1}}{N(x_{m-})},
\]
where we write $N(x_{m-})$ to mean the number of jobs in the queue at a
time infinitesimally before $x_{m}$.
Each one of the values $x_m$ is either an arrival time of some other
job, or the departure time of a job preceding $d_e$ in the departure
order. So $\partial s_i / \partial d_j = 0$ if $d_i < d_j$, and the
Jacobian matrix is again triangular. Further, $x_1 ,\ldots, x_{m-1}$ is
not a function of $d_e$, so $\partial s_e / \partial d_e =
N(d_{e-})^{-1}$. So
the joint density is
%
%
\begin{equation}
\label{eq:ps:lik}
p(\ba, \bd) = \prod_{e} N(d_{e-})^{-1} g(a_{e} - a_{e-1}) f(s_{e}).
\end{equation}

\subsection{Networks of queues}
\label{sec:network}

In this section we present a model of the paths taken by tasks through
a network
of queues. We also bring this model together with the single-queue models
discussed in previous sections and present a full probabilistic model of
networks of queues.

We begin by developing notation to describe the path of a task
through the system. For any job $e$, we denote the queue that serves
the job as $q_{e}$.
Every job has two predecessors: a within-queue predecessor $\rho(e)$,
which is the immediately previous job (from some other task) to arrive
at $q_{e}$, and a within-task predecessor $\pi(e)$, which is the
immediately previous job from the same task as $e$.
Finally, arrivals to the system as a whole are represented using
special \textit{initial jobs},
which arrive at a designated initial queue $q_0$ at time $0$ and depart
at the time that the task enters the system.
The queue $q_0$ is always single processor FCFS.
This simplifies the notation because now the interarrival times are
simply service times at the initial queue.

With this notation, a queueing network model can be defined as follows:
\begin{enumerate}[3.]
\item For every task, the path of queues is distributed according to a
Markov chain.
We denote the transition distribution of this Markov chain as $T(q' | q)$.
This is a distribution over sequences of queues, so it is a
finite-state Markov chain.
The initial state of the chain is always $q_0$. To specify $T(q' | q)$,
we assume that the modeler has defined
a network structure based on prior knowledge, such as the three-tier
structure shown in Figure~\ref{fig:3tier}.
This structure defines all of the possible successors of a queue $q$,
that is, all the queues $q'$ for which $T(q' | q)$ is nonzero;
once this structure is specified, we assume that $T(q' | q)$ is uniform
over the possible successors.
\item The arrival time for each initial job is set to zero.
\item Each service time $s_e$ is distributed independently according to
the service density for $q_e$. We will denote this density
as $f_q (s; \theta_q)$,
where $\theta_q$ are the parameters of the service distribution.
In the current work, we use exponential distributions over the service
times so that
%
%
\begin{equation}
\label{eq:2}
f_q (s; \theta_q) = \theta_q \exp\{-\theta_q s \},
\end{equation}
but our sampling algorithms are easily extended to general service
distributions.
\item The departure times are computed by solving
the system of equations that includes: (a) for every queue in the
network, the equations in
\eqref{eq:ggkfcfs:eqns}, \eqref{eq:gg1rss:eqns}, or \eqref
{eq:gg1ps:eqns}, as appropriate
(the queues in the network need not all be the same type), and
(b)~for all noninitial jobs, the equation $d_{\pi(e)} = a_e$.
We call this system of equations the \textit{departure time equations}.
\end{enumerate}
The full set of model parameters is simply the set of parameters for
all the service distributions,
that is, the full parameter vector $\theta= \{ \theta_q \mid q \in
\calN\}$, where $\calN$ is the set
of all queues in the network.

Now we present the joint density more formally.
The density can be derived in a similar fashion as that for single queues.
We write $s_e(\bd)$ to denote the service time for job $e$ that would
result from
the departure times $\bd$; this is the inverse of the transformation
defined by the departure time equations.
Because different queues are allowed to use different queueing regimes,
different jobs will have different Jacobian terms, depending on their
queue type.
Fortunately, the Jacobian matrix in a network of queues model is still
upper triangular,
as can be seen by ordering the jobs from all queues by their departure time.
This means that the joint density can still be written as a product
over jobs.

So the joint density is
%
%
\begin{equation}
p(\bd, \bq| \theta) = \prod_{e} T( q_e | q_{\pi(e)} ) h(q_e,
s_e, d_e) f (s_e(\bd); \theta_{q_e})\label{eq:lik},
\end{equation}
where the function $h$ is the queue-specific part of the likelihood:
%
%
\begin{equation}
\label{eq:1}
h(q_e, s_e, d_e) =
\cases{
1, & \quad  if $q_e$ is an FCFS queue,  \cr
N(d_e - s_e), & \quad  if $q_e$ is an RSS queue,  \cr
N(d_{e-}), & \quad  if $q_e$ is a PS queue.
}
\end{equation}
The likelihood \eqref{eq:lik} is a product over events, where for each
event $e$ there are three terms: the first involving $T$ arises from
the paths for each task, the second involving $h$ is the queue-specific
term, and the third involving $f$ arises from the service time for
event $e$.
Finally, observe that we no longer need to include terms for the
interarrival times, because of our convention
concerning initial tasks.

We will present both maximum likelihood and Bayesian approaches to
estimation. For the Bayesian approach, we need a prior on $\theta$ to
complete the model.
In this work we use the simple improper prior $p(\theta) = \prod_q
1/\theta_q$, where the product is over all queues in the network.
(This choice does mean that there is a nonzero probability in the prior
and the posterior that the system is unstable; we discuss this issue in
Section~\ref{sec:discussion}.)

To summarize, the key insight here is to view the queueing network as a
deterministic transformation from service times to departure times,
via the departure time equations.
The distinction between service times and departure times is important
statistically, because
while the service times are all i.i.d.,
the departure times have complex dependencies. For example, if $K=1$,
then the FCFS queue imposes the assumption that the arrival order is
the same as the departure order. This assumption is relaxed in the more
complex models (i.e., $K>1$ or RSS), but still some combinations of
arrivals and departures are infeasible. For example, in an RSS queue,
whenever a job $e$ arrives at a nonempty queue, at least one other job
must depart before $e$ can enter service.
In a PS queue, on the other hand, all combinations of arrivals and
departures are feasible, so long as all $a_{e} \leq d_{e}$.

 \begin{remark*}
 Because the distribution $p(\bd| \theta)$ is
high-dimensional and has special discrete structure,
it is natural to consider whether it can be interpreted as a graphical
model [\citet{lauritzenbook}].
In fact, however, as explained in Section~\ref{sec:blanket},
this distribution cannot directly be represented as a graphical model
in a useful way.
\end{remark*}

\section{Inferential problem}
\label{sec:problems}

In this section we describe the inferential setting in more detail,
and also explain several complications in queueing models that make
sampling from the posterior distribution more difficult than in
traditional multivariate models.

\subsection{Missing data}
\label{sec:observations}

First we examine the nature of the observations.
If the arrival, departure, and path information for every job were observed,
then it would be straightforward to compute the interarrival and
service times,
by inverting the departure time equations (Section~\ref{sec:network}).
Once the observations have been transformed in this way, they are independent,
so inference is straightforward.

In practice, however, complete data are not always available.
One reason for this is that the performance cost of recording detailed
data about every request
can be unacceptable in a system that receives millions of requests per day.
Another reason is that systems are built using hardware components and
software libraries from outside sources. The performance characteristics
of these outside components may not be fully understood, so there may
also be queueing effects in the system
that are unknown to the system developers; this possibility will be
discussed further in Section~\ref{sec:model-selection}.

We model the data collection process abstractly as follows.
Whenever a task arrives at the system, it is chosen for logging with
some probability $p$.
For tasks that are not chosen, which we call \textit{hidden tasks}, no
information is recorded.
For tasks that are chosen, which we call \textit{observed tasks}, the
arrival time, departure time, and queue for every job in the task are
recorded. Also, whenever a task is observed, we record the total number
of tasks, both observed and hidden, that have entered the system. This
provides additional information about the amount of workload over time,
and is
easy to collect in actual systems. Formally, the data can be described
as follows. Let $\calT$ be
the set of tasks that are chosen to be observed; each task $A$ is a set
of jobs $e \in A$. Let $d_0(A)$ be the departure time of the initial
job for task $A$. Let $N_0(t)$ be the number of tasks, both observed
and hidden, that have entered the system by time $t$. Then the data are
$O = \{ (N_0(d_0(A)), J_A) \mid A \in\calT\}$, where $J_A = \{ (a_e,
d_e, q_e) \mid e \in A \} $.

More sophisticated observation schemes are certainly possible. For
example, if the response time of the system appears to be increasing,
we may wish to collect data more often, in order to give developers
more information with which to debug the system. Or, we may use a
stratified approach, in which we collect detailed information from a
random sample of all tasks, from a random sample of tasks in the top
10th percentile of response time, and so on. We will not consider such
adaptive schemes in this work.

\subsection{Inference}

We consider both Bayesian and maximum likelihood approaches to
inference in this work.
In either case,
the key difficulty is to sample from the posterior distribution over
missing data.
In particular, two posterior distributions will be of interest.
Recall that in a queueing model, the response time $r_e$ of a job is
divided into two components as $r_e = w_e + s_e$, where the service
time $s_e$ represents the processing time that is intrinsically
required for the job, and the waiting time $w_e$ represents the
additional delay due to workload on the system.
Then the first posterior distribution of interest is the distribution
$p(\bs| O, \theta)$ over the vector of service times $\bs$ for
all jobs, hidden and observed. This distribution allows inference over
the parameters of the service distributions for each queue.
The second posterior distribution of interest is the distribution
$p(\bw| O, \theta)$ over the vector of waiting times $\bw$ for
all jobs.
This captures the fraction of the response time for each job that was
caused by workload.

Thus, our setting can be viewed as a missing data problem, in which the
missing data are the unrecorded departure times. Our goal will be to
develop an MCMC sampler for the distribution $p(\bd| O, \theta)$
over departure times. Once we have samples from $p(\bd| O, \theta
)$, we can obtain samples from $p(\bs| O, \theta)$ and $p(\bw
| O, \theta)$ by inverting the departure time equations.
Furthermore, once we have a sampler for $p(\bd| O, \theta)$,
parameter estimation is also straightforward. In a Bayesian approach,
we simply alternate the sampler for $p(\bd| O, \theta)$ with a
Gibbs step for $p(\theta| \bd)$.
In a maximum likelihood approach, we use stochastic EM [\citet{celeux85sem}].

However, designing an efficient sampler for $p(\bd| O, \theta)$
is complex, because the latent variables have many complex dependencies.
In the next two subsections we describe these difficulties in more detail,
highlighting their effect on the algorithm that we will eventually propose.

\subsection{Difficulties in proposal functions}
\label{sec:proposal}
\label{sec:singularities}

A natural idea is to sample from the posterior distribution over the
missing data using either an importance sampler, a rejection sampler,
or a Metropolis--Hastings algorithm.
But designing a good proposal is difficult for even the simplest
queueing models,
because the shape of
the conditional distribution varies with the arrival rate.
To see this, consider two independent single-processor FCFS queues,
each with three arrivals, as shown below:

\begin{figure}[h]

\includegraphics{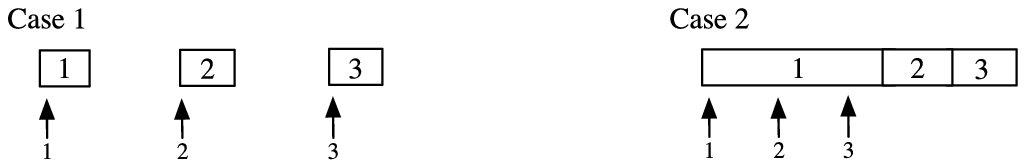}

\end{figure}

Here the horizontal axis represents time, the vertical
arrows indicate when jobs arrive at the system, and the boxes
represent the intervals between when jobs enter service and when they
finish, that is, the service
times.
The interarrival distribution is exponential with rate $\lambda$, and
the service distribution is exponential with rate $\mu$.

For each of these two queues, suppose that we wish to resample the
arrival time of job 2, conditioned on the rest of the system state, as
we might wish to do within a Gibbs sampler.
In case 1, the queue is lightly loaded ($\lambda\ll\mu$), so the
dominant component of the response time is the service time. Therefore,
the distribution $a_{2} = d_{2} - \Exp{\mu}$ is an excellent proposal
for an importance sampler. (It is inexact because the shape of the
distribution changes in the area $a_{2} < d_{1}$.)
In case 2, however, this proposal would be extremely poor, because in
this heavily loaded case,
the true conditional distribution is $\operatorname{Unif}[ a_{1}
; a_{3} ]$.
A better proposal would be flat until the previous job departs and then
decays exponentially.
But this is precisely the behavior of the exact conditional
distribution, so we consider that instead.

\subsection{Difficulties caused by long-range dependencies}
\label{sec:blanket}

In this section we describe another difficulty in queueing models:
the unobserved arrival and departure times have complex dependencies.
Namely, modifying the departure time of one job can force modification
of service times of jobs
that occur much later, if all other arrival and departure times are
kept constant.
In the terminology of graphical modeling [\citet{lauritzenbook}],
this means that the Markov blanket of a single departure can
be arbitrarily large.

\begin{figure}

\includegraphics{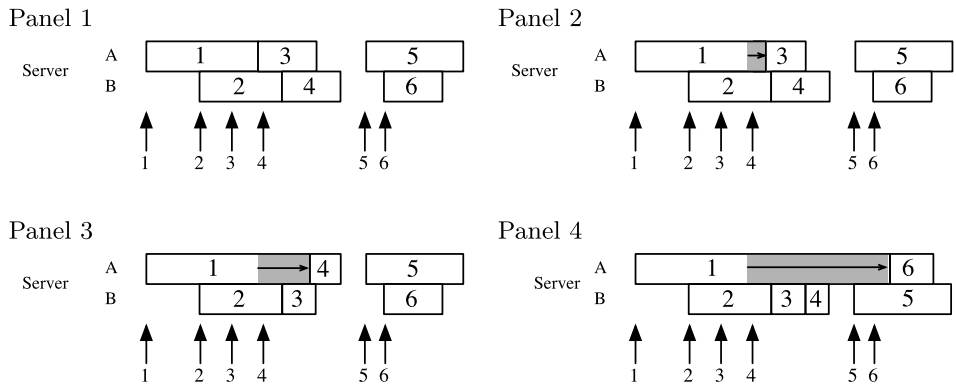}

\caption{A departure with a large Markov blanket.}
\label{fig:blanket}
\end{figure}

This can be illustrated by a simple example.
Consider the two-processor FCFS queue shown in Figure~\ref{fig:blanket}.
Panel 1 depicts the initial state of the sampler,
from which we wish to resample the departure $d_{1}$ to a new
value $d_{1}'$, holding all departures
constant, as we would in a Gibbs sampler, for example. Thus, as $d_{1}$
changes, so will the service times of jobs 3--6.

%

Three different choices for $d_{1}'$ are illustrated in panels 2--4 of
Figure~\ref{fig:blanket}.
First, suppose that $d_{1}'$ falls within the range $(d_{1}, d_{2})$
(second panel). This has the effect of
shortening the service time $s_{3}$ without affecting any other jobs.
If instead $d_{1}'$ falls in $(d_{2}, d_{4})$ (third panel), then both
jobs 3 and 4 are affected:
job 3 moves to server $B$, changing its service time; and job 4 enters
service immediately after job~1 leaves.
Third, if $d_{1}'$ falls even later, in $(a_{6}, d_{6})$ (fourth
panel), then both jobs~3 and 4
move to server B, changing their service times; job 5 switches
processors but is otherwise unaffected; and job 6 can start only when
job 1 leaves.
Finally, notice that it is impossible for $d_{1}'$ to occur later than
$d_{6}$ if all other departures are held constant. This is because job
6 cannot depart until all but one
of the earlier jobs depart, that is, $d_{6} \geq\min[ d_{1}',
d_{5} ]$.
So since $d_{5} > d_{6}$, it must be that $d_{6} \geq d_{1}'$.

This phenomenon complicates the development of a sampler because of the
difficulty that it creates in computing the conditional distributions
required by a Gibbs sampler, particularly in computing their
normalizing constants. In the previous example, for instance, the
conditional distribution over $d_{1}$ cannot in general be computed in
closed form. But numerical integration of the unnormalized density is
also difficult, because the density has singularities at the times when
other jobs arrive and depart, for example, at times $d_2$, $d_4$, and
$a_6$ above. Furthermore, even without the normalizing constant,
computing the density at some point $d_1'$ requires computing new
values for the service times of the succeeding jobs. If the new value
$d_1'$ affects many subsequent jobs, then the computational cost
required to compute the conditional density will be large.

Furthermore, this point has important consequences from a
representational perspective.
It is natural to suspect that the distribution over arrival times,
departure times, and auxiliary variables
could be represented as a directed graphical model. In fact, however,
because the Markov blanket for each
departure time is unbounded, the distribution over departure times
cannot be represented as a graph in a useful way.

This may seem to present a severe difficulty, because sampling
algorithms for high-dimensional multivariate distributions rely on the
Markov blankets being small in order to achieve computational efficiency.
Fortunately, even though the Markov blankets can be large, the
``expected Markov blankets'' are often small, by which we mean that
these long-range effects occur only for large values of the departure
times that are unlikely in the posterior. We expect that typical values
of departure times will be smaller and will therefore have only local
effects on the queue.
This situation will be sufficient to allow us to develop a
computationally efficient sampler.

Finally, note that the phenomenon in this example occurs even if the
queue is Markovian, that is, if the interarrival and service times are
exponentially distributed. Such queues are called Markovian because the
process that governs the number of jobs in queue is a continuous-time
Markov chain. However, the sequence of departure times $(d_1, d_2,
\ldots, d_n)$ is \textit{not} in general a discrete-time Markov chain,
except in the special case in which the queue contains only one
processor. This is one reason that analysis in queueing theory often
focuses on the number-in-queue representation, but
in our setting, the data contain arrival and departure times, and it is
not possible to translate directly between the two representations when
the data are incomplete.

\section{Sampling}
\label{sec:inference}

In this section we describe the details of a sampler that
addresses the difficulties discussed in the previous section.
We focus on the sampler for the posterior
$p(\bd| O)$ over the vector $\bd$ of all departure times.
Once we have samples from this distribution,
we can easily obtain samples of service times and waiting times by
inverting the
departure time equations, once for each of the departure samples.
Furthermore, inference about the parameters can be performed in the
usual fashion
using either a Gibbs step over the model parameters in a Bayesian framework,
or using Monte Carlo EM in a maximum likelihood framework.
Exact sampling from the posterior $p(\bd| O)$ is infeasible
even for the simplest queueing models, so instead we sample approximately
using Markov chain Monte Carlo (MCMC).

Our sampler is an instance of a slice sampler [\citet
{damien99gibbs}; \citet{Neal2003Slice}].
We recall briefly the setup for slice sampling.
Suppose that we wish to sample from a distribution with density
$p(x)$, where $x$ is real valued. This can be accomplished by sampling
uniformly from the two-dimensional region under $p$, that is, the
region $R = \{ (x,u) \mid 0 < u < p(x) \}$, because this distribution
has marginal $p(x)$.
The slice sampler is essentially a coordinatewise Gibbs update that
leaves the uniform distribution on $R$ invariant.
In the simplest version, given a current iterate $(x,u)$, the sampler
alternates between (a) ``vertically'' sampling a new value $u'$ from
the uniform distribution on $(0, p(x))$, and (b) ``horizontally''
sampling a new value $x'$ uniformly from the so-called \textit
{horizontal slice}, that is, the set of points $(x', u')$ where $(x',
u') \in R$ and also $u' = u$.
Both of these updates leave the uniform distribution over $R$ invariant.
In practice, the horizontal slice cannot be computed exactly, but \citet
{Neal2003Slice}
discusses several other horizontal updates in the same spirit that are
easy to compute.
For multivariate $x$, the slice sampler can be applied in turn to each
component.

As described in the previous section, certain difficulties in queueing models
make it difficult to apply simple Gibbs or Metropolis--Hastings samplers.
The slice sampler circumvents these difficulties, because it
requires only the ability to compute
the unnormalized conditional density, not the ability to sample from it
or to compute its normalizing constant.
The following sections describe how we compute
the unnormalized conditional density.

\subsection{Overview}
\label{sec:overview}

In each update of the sampler, we sample a new value of the departure
time $d_{e}$ for some job $e$,
using a slice sampling kernel with stationary distribution $p(d_e | \bd
_{\backslash e})$,
where $\bd_{\backslash e}$ refers to the vector of all departure times
but $d_e$.
Because the slice sampler only requires the density up to a constant,
it is sufficient to compute the joint $p(\bd)$.

The joint density can be computed pointwise for a given $\bd$ by
inverting the departure time equations to obtain the corresponding set of
service times, and then using \eqref{eq:lik}. This equation contains a
product over all $N$ jobs.
Computing this product naively at every update of the sampler would
require $O(N)$ time, so that a full pass through the data
would require $O(N^{2})$ time. This quadratic computational cost is
unacceptable for the large numbers of
jobs that can be generated by a real system. Fortunately, this cost can
be circumvented using a lazy updating scheme, in which first
we generate the set of \textit{relevant jobs} $\Delta$ that would be
changed if the new value of $d_{e}$ were to be adopted.
Then we incrementally update the factors in the product \eqref{eq:lik}
only for the relevant jobs.
\comment{Then the new log density can be incrementally computed as
\[
\ell^{\new} = \ell^{\old} + \sum_{e \in\Delta} \log f(s^{\new
}_{e}) - \log f(s^{\old}_{e}) + \log h(\NQ{e}^{new})- \log h(\NQ{e}^{old}),
\]
where $\ell^{\old}$, $s^{\old}$, and $\NQ{e}^{\old}$ are stored
values from the previous iteration of the sampler.
If any $s^{\new}_{e}$ is negative, this means that the value $a_{e} =
d_{\pi(e)}$ selected by the slice sampler is infeasible, so set $\ell
^{\new} = -\infty$.}

Essentially, computing the unnormalized
density requires that we compute the list of jobs whose service time would
be affected if a single departure time changed.
This amounts to setting $d_{e}$ to the new value,
propagating these two changes through the departure time equations, and
obtaining a
new service time $s_{e'}$ for all other jobs in the two queues $q_{e}$
and $q_{\pi^{-1}(e)}$.

\begin{algorithm}
\caption{Update the service times for a departure change in a
$K$-processor FCFS queue}\label{alg:ggk:departure}
\begin{algorithmic}[1]
\STATE\textbf{function} \FN{UpdateForDeparture}($e_{0}$)
\STATE\Comment{Input: $e_{0}$, job with changed departure}
\STATE\VAR{stabilized} $\gets$ 0
\STATE$e \gets$ $\rho^{-1}(e_{0})$
\WHILE{$e \neq$ \NULL\ and not \VAR{stabilized}}
\STATE$b_{ek} \gets b_{\rho(e),k} \qquad\forall k \in[0,K)$
\STATE$b_{e,k(\rho(e))} \gets d_{\rho(e)}$
\STATE$\VAR{stabilized} \gets1 \textbf{ if } b_{e} = \mbox{ old
value of } b_{e} \textbf{ else } 0$
\STATE$c_{e} \gets\min_{k \in[0,K]} b_{ek}$
\STATE$p_{e} \gets\arg\min_{k \in[0,K]} b_{ek}$
\STATE$s_e \gets d_{e} - \max[ a_{e}, c_{e} ]$
\STATE$e \gets\rho^{-1}(e)$
\ENDWHILE
\end{algorithmic}
\end{algorithm}

\begin{algorithm}
\caption{Update the service times for an arrival change in a
$K$-processor FCFS queue}\label{alg:ggk:arrival}
\begin{algorithmic}[1]
\STATE\textbf{function} \FN{UpdateForArrival}($e_{0}$, \VAR{aOld})
\STATE\Comment{Input: $e_{0}$, job with changed arrival}
\STATE\Comment{Input: \VAR{aOld}, old arrival of job $e_{0}$}
\STATE\Comment{Update arrival order $\rho$ due to $e_{0}$}
\STATE$\VAR{aMin} \gets\min[ a_{e_{0}}, \VAR{aOld} ]$
\STATE$\VAR{aMax} \gets\max[ a_{e_{0}}, \VAR{aOld} ]$
\STATE$E \gets\mbox{all jobs arriving within \VAR{aMin} \ldots\VAR{aMax}}$
\STATE\Comment{First change jobs that arrive near $e_{0}$}
\FORALL{$e \in E$}
\STATE$b_{ek} \gets b_{\rho(e),k} \qquad\forall k \in[0,K)$
\STATE$b_{e,k(\rho(e))} \gets d_{\rho(e)}$
\STATE$c_{e} \gets\min_{k \in[0,K]} b_{ek}$
\STATE$p_{e} \gets\arg\min_{k \in[0,K]} b_{ek}$
\STATE$s_e \gets d_{e} - \max[ a_{e}, c_{e} ]$
\ENDFOR
\STATE\Comment{Second, propagate changes to later jobs}
\STATE$e \gets\rho^{-1}(\FN{lastElement}(E))$
\STATE$\VAR{stabilized} \gets1 \textbf{ if } b_{e} = \mbox{old
value of } b_{e} \textbf{ else } 0$
\IF{not \VAR{stabilized}}
\STATE\FN{UpdateForDeparture}($e$)
\ENDIF
\end{algorithmic}
\end{algorithm}

\begin{algorithm}
\caption{Update the service times for a departure change in an RSS queue.}
\label{alg:ggrss:departure}
\begin{algorithmic}[1]
\STATE Update departure order $\gamma$ for changed departure $d_{e}$
\STATE\VAR{newPrev}, \VAR{newNext} $\gets$ Jobs departing
immediately before and after the time $d_{e}^{\old}$
\STATE\VAR{oldPrev}, \VAR{oldNext} $\gets$ Jobs departing
immediately before and after the time $d_{e}$
\STATE$\VAR{dMin} \gets\min[ d_{\mathit{newPrev}}, d_{\mathit
{oldPrev}} ]$
\STATE$\VAR{dMax} \gets\max[ d_{\mathit{newNext}}, d_{\mathit
{oldNext}} ]$
\STATE$L \gets$ all jobs with departures in \VAR{dMin} \ldots\VAR{dMax}
\FORALL{$e \in L$}
\STATE$u_{e} \gets\max[ a_{e}, d_{\gamma(e)} ]$
\STATE$s_{e} \gets d_{e} - u_{e}$
\ENDFOR
\end{algorithmic}
\end{algorithm}

So for each type of queue, we require two algorithms: (a) a \textit
{propagation algorithm} that
computes the modified set of service times that results from a new
value of $d_{e}$, and (b) a \textit{relevant job set} algorithm that
computes the set of jobs $\Delta$ for which the associated factor in
\eqref{eq:lik} needs to be updated. It is not the case in general that
$\Delta$ is just the set of jobs whose service times are changed by
the propagation algorithm; this is because of the factor $h(q_e, s_e,
d_e)$ in \eqref{eq:lik}.
The next three sections describe the propagation and relevant job set
algorithms for FCFS queues (Section~\ref{sec:sampler:fcfs}), RSS
queues (Section~\ref{sec:sampler:rss}), and PS queues (Section~\ref
{sec:ggps}).

\subsection{FCFS queues}
\label{sec:sampler:fcfs}

The propagation algorithms for the FCFS queue are given in
Algorithm~\ref{alg:ggk:departure} (for the departure times)
and Algorithm~\ref{alg:ggk:arrival} (for the arrival times). These
algorithms compute new values
of $b_{e'k}$, $u_{e'}$, $c_{e'}$, $p_{e'}$, and $s_{e'}$ for all other
jobs $e'$ and processors $k$ for all other jobs in the queue.
The main idea is that any service time $s_{e'}$ depends on its previous
jobs only through the processor-clear times $b_{\rho(e')k}$ of the
immediately previous job $\rho(e')$. Furthermore, each $b_{ek}$ can be
computed recursively as $b_{ek} = d_{\rho(e)}$ if $k = p_{\rho(e)}$
and $b_{ek} = b_{\rho(e),k}$ otherwise.

A separate relevant job set algorithm is unnecessary for the FCFS
queue. Because for this queue $h(q_e, s_e, d_e) = 1$,
the relevant job set is simply the set of jobs whose service times are updated
by Algorithms~\ref{alg:ggk:departure} and \ref{alg:ggk:arrival}.

\subsection{RSS queues}
\label{sec:sampler:rss}

The propagation algorithm for an RSS queue is given in Algorithm~\ref
{alg:ggrss:departure}.
This algorithm is used for departure changes.
For an arrival change, on the other hand, none of the service times for
other jobs in $q_{e}$ need to be updated.

Two algorithmic issues are specific to RSS queues.
First, the new value $a_{e} = d_{\pi(e)}$ must
still be feasible with respect to the constraints \eqref{eq:gg1rss:eqns}.
This can be ensured by computing the new departure order $\gamma$ for
$q_{\pi(e)}$,
and then verifying for all jobs in $q_{e}$ and $q_{\pi(e)}$
that $\gamma^{-1}(e) \in\IQ{e}$. If this is not the case, then the
potential new value $a_{e} = d_{\pi(e)}$
is rejected.

Second, observe from \eqref{eq:1} that
$h(q_e, s_e, d_e) = \NQ{e}^{-1}$. (Recall that the commencement time
$u_e = d_e - s_e$ is the time that $e$ enters service.) These factors
arise from the random selection of
job $e$ to enter service, out of the $\NQ{e}$ jobs that were in queue.
Intuitively, these factors are the only penalty on a job waiting in
queue for a long time;
without them, the sampled waiting times would become arbitrarily
large.
To compute these, we need an efficient data structure for computing
$\NQ{e}$,
the number of jobs in queue when the job $e$ entered
service. This is implemented by two sorted lists for each queue:
one that contains all of the queue's arrival times, and one that
contains all of the departure times. Then we use a binary
search to compute the total number of jobs that have arrived before
$u_{e}$ (denoted as $\#A_{e}$) and the total number of jobs that have
departed before $u_{e}$ (denoted as $\#D_{e}$). Then we can compute
$\NQ{e} = \#A_{e} - \#D_{e}$.

\begin{algorithm}
\caption{Update dependent service times for an arrival or a departure
change in a PS queue.}
\label{alg:ggps}
\begin{algorithmic}[1]
\STATE\textbf{function} \FN{UpdateJobs}$(e, \VAR{aOld}, \VAR{dOld})$
\STATE\Comment{Update dependent jobs for an arrival or a departure
change to the job $e$}
\STATE\Comment{Input: $e$, job with changed arrival or departure}
\STATE\Comment{Input: $\VAR{aOld}, \VAR{dOld}$, old arrival and
departure times of $e$}
\STATE Recompute $N(t)$ for new arrival and departure times of $e$
\STATE$\Delta\gets\FN{RelevantJobs}(e, \VAR{aOld}, \VAR{dOld})$
\FORALL{$e' \in\Delta$}
\STATE$\displaystyle s_{e'} \gets\int_{a_{e'}}^{d_{e'}} \frac
{1}{N(t)}\,dt$
\ENDFOR

\STATE\textbf{function} \FN{RelevantJobs}$(e, \VAR{aOld}, \VAR{dOld})$
\STATE\Comment{Compute set of jobs that are affected by change to the
job $e$}
\STATE\Comment{Input: $e$, job with changed arrival or departure}
\STATE\Comment{Input: $\VAR{aOld}, \VAR{dOld}$, old arrival and
departure times of $e$}
\STATE$a \gets\min[ a_e, \VAR{aOld} ]$
\STATE$d \gets\max[ d_e, \VAR{dOld} ]$
\RETURN$\{ e' | \mbox{$(a_{e'}, d_{e'})$ intersects $(a, d)$}\}$
\end{algorithmic}
\end{algorithm}

Finally, the set of relevant jobs $\Delta$ must include all jobs $e'$
whose commencement time
falls in $(a_{e}^{\old}, a_{e}^{\new})$, because those jobs will have
a new value of $\NQ{e'}$.
This set can be computed efficiently using a data structure that
indexes jobs by their commencement time.

\subsection{PS queues}
\label{sec:ggps}

The propagation algorithm for the PS queue is given by the function \FN
{UpdateJobs} in Algorithm~\ref{alg:ggps}. The same algorithm is used
for arrival and departure changes.
This algorithm computes new service times directly by solving the
relevant departure time equations \eqref{eq:gg1ps:eqns} for the
service times, with the new departure
times fixed.
For the PS queue, $h(q_e, s_e, d_e) = N(d_{e-})$, so again an efficient
data structure is required to compute
the step function $N(t)$, the number of jobs in the queue at time $t$.
The same data structure is used as in the RSS queue.

The relevant job set algorithm for PS queues is given by the function
\FN{RelevantJobs} in Algorithm~\ref{alg:ggps}. The idea here
is that when a departure time changes from $d_e$ to $d_{e}'$, all jobs
that are in the system during any portion of that interval will have
their service times affected, because of the change to $N(t)$. So
computing the set of relevant jobs amounts to searching through a set
of intervals to find all those that intersect $(d_e, d_{e'})$.
Efficient data structures are required here as well;
in our implementation, we use a variation of a treap [\citet{clrs}]
designed to store intervals.

\subsection{Initialization}
\label{sec:initialization}

A final issue is how the sampler is initialized. This is challenging
because not all sets of arrivals and departures are feasible: the departure
time equations define a set of nonconvex constraints.
In addition,
the initial configuration should also be suitable for mixing. For
example, setting all latent interarrival and service
times to zero results in a feasible configuration, but\vadjust{\goodbreak} one that
makes mixing difficult. Or, if the service
distribution belongs to a scale family (such as gamma or lognormal),
initializing the service times
to nearly identical values causes the initial variance to be sampled to
a very
small value, which is also bad for mixing.

\comment{
For a network of RSS queues in which the initial arrival and
final departure time of the task are known, we can get a reasonably
good initialization by using a linear program to find a feasible point
of the constraints $a_{e} \leq d_{e}$; $d_{\pi(e)} = a_{e}$ that is consistent
with the observations. Given a
set of arrival and departure times, we can compute the resulting
service times deterministically. Finally, we run a ``sweetening step''
in which we run a Gibbs sampler for a few iterations with exponential
service distributions before switching to the actual service
distributions used in the model. This ensures that the service times
are not all equal, avoiding the zero-variance problem mentioned above.

More generally, for $K$-processor FCFS queues, or when a subset of
tasks is
unobserved, a more complex initialization procedure is necessary.}
Initialization proceeds as follows. For
each unobserved task, we sample a path of queues from the prior
distribution over paths, and
service times from an exponential distribution initialized from the mean
of the observed response times. Sometimes the sampled service time
will conflict with the observed arrivals and departures. In this case
we use rejection, and if no valid service time can be found, we set
the service time to zero. Finally, we run a few Gibbs steps
with exponential service distributions, before switching to the actual
service distributions in the model. This prevents zero service
times, which would cause zero-likelihood problems with some service
distributions, such as the lognormal.

\section{Experimental setup}
\label{sec:expt-setup}

In this section we describe the Web application whose performance we
analyze in
the remainder of the paper.
Cloudstone [\citet{Cloudstone}] is an application that has been proposed
as an experimental setup for academic study of the behavior of Web 2.0
applications
such as Facebook and MySpace.
The Cloudstone application was developed by a professional Web
developer with the intention of reflecting common
programming idioms that are used in actual applications.
For example, the version of Cloudstone that we use is implemented in
Ruby on Rails,
a popular software library for Web applications that has been
used by several high-profile commercial applications, including
Basecamp and Twitter.

%
\begin{figure}[b]

\includegraphics{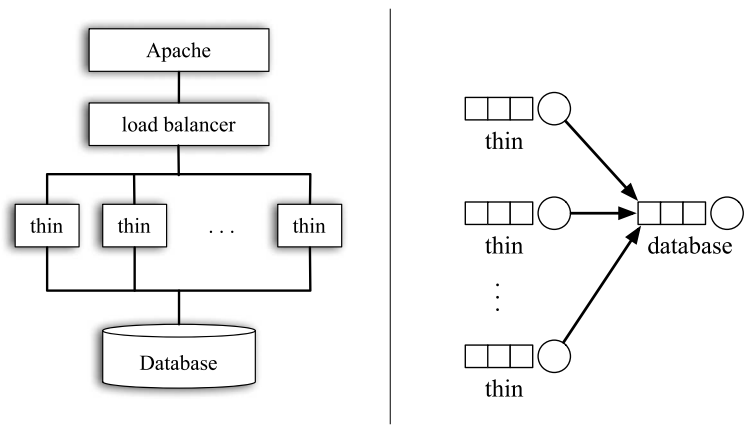}

\caption{Architecture of the Cloudstone Web application \textup{(left)}. Figure
adapted from \protect\citet{Cloudstone}.  Queueing model of Cloudstone
application \textup{(right)}.}
\label{fig:cloudstone-architecture}
\label{fig:cloudstone-model}
\end{figure}

The architecture of the system follows common practice for medium-scale Web
services, and is shown in Figure~\ref{fig:cloudstone-architecture}.
Incoming requests arrive first at \swname{apache}, a~popular Web server.
In this system, \swname{apache} is used to serve ``static content,''
that is,
information that does not need to be recomputed for each user,
such as images, the help pages, and so on.
If the external request asks only for static content, then \swname
{apache} handles
the response directly, and no processing is required by the rest of the system.
``Dynamic content,'' on the other hand,
comprises pages that need to be computed separately for each user of
the system,
such as a user's email inbox, or a user's list of contacts on a social
networking site.
Dynamic content changes over time, so that if the same user makes the
same request at different times,
the correct response may well be different, so the response needs to be
computed afresh.

Requests for dynamic content are handled by a Web server called \swname
{thin}, which is specially designed to run the Ruby on Rails library.
In order to handle a large volume of requests, multiple copies of
\swname{thin} are run on separate machines. The copies of \swname
{thin} do not themselves store any information on individual users, so
they are equivalent in their ability to handle external requests.
Requests are distributed among the \swname{thin}s by a piece of
software called a \textit{load balancer}, whose sole purpose is to
rapidly assign requests at random to one of a set of equivalent copies
of a service. Because they do not store user data, the copies of
\swname{thin} need some mechanism for obtaining this data from
elsewhere. This is handled by having a database running on a separate
machine, with which all of the copies of \swname{thin} communicate.
In our setup, we run 10 copies of \swname{thin} on 5 machines, two
copies per machine.
We run the \swname{apache} server, the load balancer, and the database
each on their own machine,
so that the system involves 8 machines in all.

We run a series of 2663 requests to Cloudstone over 450~s, using the
workload generator included with the benchmark. A total of 7989 jobs
are caused by the 2663 requests.
The workload is increased steadily over the period, ranging from 1.6
requests$/$second at the beginning to 11.2 requests$/$second at the end.
The application is run on Amazon's EC2 utility computing service.
For each request, we record which of the \swname{thin}s handled the request,
the amount of time spent by the Rails library, and the amount of time
spent in the database. Each Cloudstone request causes many database
queries; the time we record is the sum of the time for those queries.

\section{Prediction}

In this section we demonstrate that networks of queues can effectively
extrapolate from the performance of the system at low workload to the
worse performance that occurs at higher workload. This prediction
problem is of practical importance because if the performance
degradation can be predicted in advance, then the system's developers
can take corrective action.

We compare the prediction error of a variety of queueing models on the
Cloudstone data described in Section~\ref{sec:expt-setup}. To measure
the extrapolation error, we estimate model parameters during low
workload---the first 100~s of the Cloudstone data---and evaluate the
models' predictions under high workload---the final 100~s of the data.
The workload during the training regime is 0.9 requests$/$second, whereas
the workload in the prediction regime is 9.8 requests$/$second.
During the training period, the average response time is 182 ms, while
during the prediction period the average response time is 307 ms.
The goal is to predict the mean response time over 5 second intervals
during the prediction period, given the number of tasks that arrive in
the interval.

We evaluate several queueing models: (a) single-processor RSS, (b) a
network of RSS queues, (c) a single 3-processor FCFS queue, and (d) a
network of PS queues. The networks of queues use the structure shown in
Figure~\ref{fig:cloudstone-architecture}. In all cases, the service
distributions are exponential.
For the single-queue models, the data consists of the arrival and
departure of each task from the system as a whole. For the network
models, the data consists of all arrivals and departures to the \swname
{thin}s and the database servers.
As baselines, we consider several regression models: (a)~a linear
regression of mean response time onto workload, (b) a regression that
includes linear and quadratic terms, and (c) a ``power law'' model,
that is, a linear regression of log response time onto log workload.
In all cases, the data contains information about all tasks in the
training period, that is, there is no missing data, so parameter
estimation is done by simple maximum likelihood.

%
\begin{table}
\tabcolsep=0pt
\tablewidth=265pt
\caption{Extrapolation error of performance models of Cloudstone. We
report root mean squared error on the prediction of the response time
under high workload, when training was performed under low
workload}\label{tbl:extrapolate}
\begin{tabular*}{265pt}{@{\extracolsep{\fill}}lcd{4.0}@{}}
\hline
&& \multicolumn{1}{c@{}}{{\textbf{RMSE (ms)}}} \\
\hline
&Linear regression& 258   \\
&Quadratic regression& 250   \\
&Power law regression& 194   \\
\hline
Single queue & 1-processor RSS & 1340   \\
Network & 1-processor RSS & 168   \\
Single queue & 3-processor FCFS & \multicolumn{1}{c@{}}{\phantom{000}$\mathbf{71.7}$}   \\
Network & PS & 234   \\
\hline
\end{tabular*}
\end{table}

The prediction error for all models is shown in Table~\ref{tbl:extrapolate}.
The best queueing model extrapolates markedly better than the best
regression model, with a 63\% reduction in error.
Interestingly, different queueing models extrapolate very differently,
primarily because they make different assumptions about the system's capacity.
This point is especially important because previous work on statistical
inference in queueing models has considered
only the simplest types of queueing disciplines, such as 1-processor
FCFS. These results show that the more complex
models are necessary for real systems.

A second difference between the regression models and the queueing
model is in the types of errors they make.
When the regression models perform poorly, visual inspection suggests
that noise in the data
has caused the model to oscillate wildly outside the training data
(e.g., to make negative predictions).
When the queueing models perform poorly, it is typically because the
model underestimates the capacity of the system,
so that the predicted response time explodes at a lower workload than
the actual response time.


\section{Diagnosis}
\label{sec:diagnosis}

In this section we demonstrate that our sampler can effectively
reconstruct the missing arrival and departure data.
The task is to determine which component of Cloudstone (\swname{thin}
or the database) contributes
most to the system's total response time, and how much of the response
time of that component
is due to workload. Although we
measure directly how much time is spent by Rails and by the database,
the measurements do not indicate how much of that time is
due to intrinsic processing and how much is due to workload. This
distinction is important in practice: If system response time is due to
workload, then we expect adding more servers to help, but not if
it is due to intrinsic processing.
Furthermore, we wish to log departure times from as few tasks as
possible, to minimize the logging overhead on the \swname{thin}s.

More specifically, our goal will be to estimate $\bs$, the service
times for all jobs, given an incomplete sample of arrivals and
departures. The model parameters are unknown, so those must be
estimated as well.
The data are collected using the observation scheme described in
Section~\ref{sec:observations}. We compare the estimates of $\bs$
when information from 25\%, 50\%  and 100\% of the tasks is used in the
analysis.

We model Cloudstone by a network of PS queues (Figure~\ref
{fig:cloudstone-model}): one for each \swname{thin} (10 queues in all)
and one for the database. The delay caused by \swname{apache}, by the
load balancer, and the internal network connection
is minimal, so we do not model it. The service distributions are
exponential. Parameter estimation is performed in a maximum likelihood
framework using stochastic EM, in which the missing data are imputed
using the sampler described in Section~\ref{sec:inference}.

%
\begin{figure}[b]

\includegraphics{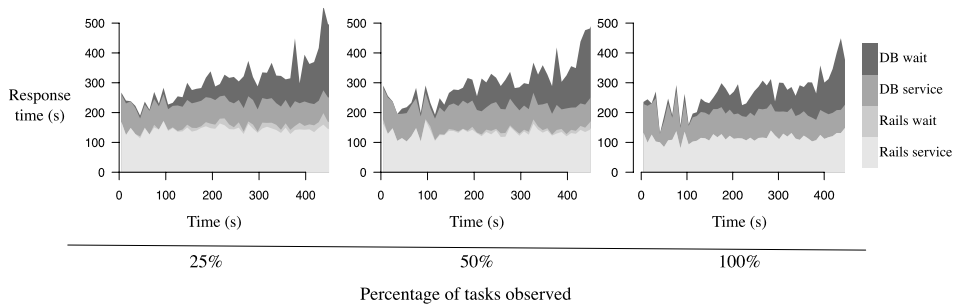}

\caption{Reconstruction of the percentage of request time spent
in each tier, from 25\% tasks observed \textup{(left)}, 50\% tasks observed (\textup{center}),
and all tasks observed (\textup{right}). The x-axis is the time in seconds that the
task entered the system, and the y-axis is the estimated service and
waiting time.}
\label{fig:reconstruction}
\end{figure}

Figure~\ref{fig:reconstruction} displays the proportion of time
per-tier spent in processing and in queue, as estimated using the slice
sampler from 25\%, 50\%  and 100\% of the full data.
Visually, the reconstruction from only 25\% of the data strongly
resembles the full data: it is apparent that as the workload increases
from left to right, the \swname{thin}s are only lightly loaded, and
the increase in response time is due to workload on the database tier.

To obtain a quantitative measure of error, we partition time into 50
equal-sized bins, and compute the mean service time for each bin and
each tier of the system.
We report the root mean squared error (RMSE) between the reconstructed
service times from the incomplete data and the service times that would
have been inferred had the full data been available. We perform
reconstruction on ten different random subsets of 25\% and 50\% of the jobs.
We use two baselines:
(a) one that always predicts that the response time is composed only of
the service time (denoted ``Wait${}={}$0'') and (b)
a linear regression of the per-job waiting time onto
the workload in the last 500 ms. Results are reported in Table~\ref
{tbl:diagnosis}.

%
\begin{table}
\tabcolsep=0pt
\tablewidth=300pt
\caption{Error at determining service times. The error measure shown
is the root mean squared error on the predicting of service times in
the full data.
The small numbers indicate the standard deviation over ten repetitions
with different observed jobs}
\label{tbl:diagnosis}
\begin{tabular*}{300pt}{@{\extracolsep{\fill}}lcc@{}}
\hline
& \multicolumn{2}{c@{}}{\textbf{RMSE (ms)}} \\[-5pt]
& \multicolumn{2}{c@{}}{\hrulefill}\\
&  $\bolds{25\%}$  &  $\bolds{50\%}$  \\
\hline
Wait $=$ 0 & \multicolumn{2}{c@{}}{$62.3$} \\
[3pt]
Linear regression & $80.4 \pm1.0 $   & $80.2 \pm0.8 $   \\
Network of queues (PS) & $50.0 \pm3.5 $   & $28.5 \pm3.3 $\\
\hline
\end{tabular*}
\end{table}

The posterior sampler performs significantly better at reconstruction
than the baselines, achieving a 25\%
reduction in error for 25\% data observed, and a 54\% reduction in
error for 50\%
data observed. Linear regression performs poorly on this task,
performing worse than
the trivial ``Wait${}={}$0'' baseline. Interestingly, the performance of
linear regression, unlike the queueing network,
does not improve with additional data. This supports the idea that the
poor performance of linear regression
is due to limitations in the model.

\section{Model selection}
\label{sec:model-selection}

A final application of our framework is model selection. Although model
selection has received relatively little attention in the context of
queueing models, it has the potential to be greatly useful, because the
performance characteristics of a software system are often not
completely understood even to its developers. For example, often
systems are built from external components, such as software libraries,
whose internal workings are not fully known. Furthermore, even if the
source code for every component is available,
system performance may differ from expectations because of
software bugs, hardware failures, or misconfiguration of the system. In
either case, a concise model can serve as a summary of system
performance under different workloads, revealing queueing dynamics that
may be unexpected.

%
\begin{figure}

\includegraphics{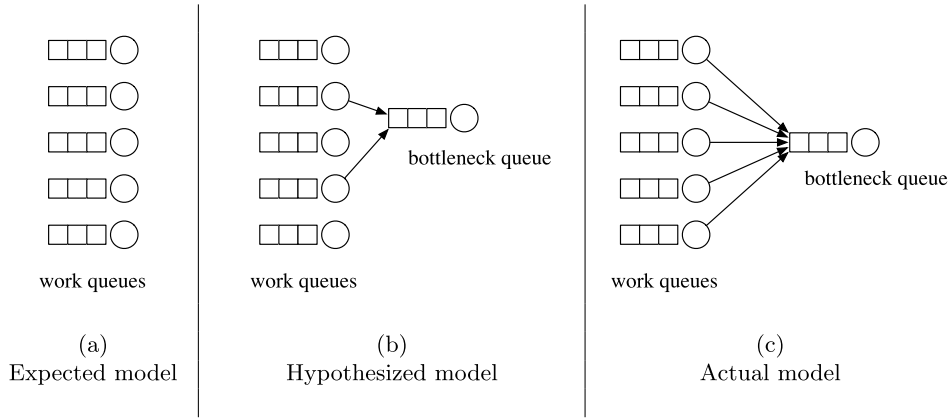}

\caption{Alternative models for missing-queue detection. At left,
queueing network representing the null model. Center, network
representing a hypothetical alternative model. At right, model used to
generate synthetic data for the experiments described in
Section~\protect\ref{sec:model-selection}.}
\label{fig:alternatives}
\end{figure}


We demonstrate this idea on a task that we call \textit{missing queue
detection}.
Suppose that we are analyzing a system whose expected behavior is
described by the queueing network in Figure~\ref{fig:alternatives}(a),
which consists of a pool of independent ``work queues.'' If there is a
bug in the system, however, the actual performance behavior may be
different than expected. If there is a bug, we might hypothesize that
the actual system behavior is described by Figure~\ref
{fig:alternatives}(b), in which a ``bottleneck queue'' has been added,
which has the effect of coupling the response times of all tasks that
are served by different work queues.
The data consists of the arrival and departure time of each task from
the system as a whole, and the identity of the ``work queue'' to which
the task was assigned. The transition time between the work queues and
the bottleneck queue is necessarily unobserved.
On the basis of such data, the goal of missing queue detection is to
choose between the models in
Figure~\ref{fig:alternatives}(a) and~(b), in other words, to determine whether the system
exhibits unexpected queueing dynamics.

A natural approach is based on least angle regression (LARS) [\citet
{efron04lars}]. In this approach, we perform a regression of the
average response time of jobs at each of the work queues, aggregated
over 5-second intervals, onto two covariates: the number of jobs at
that work queue, and the number of jobs arriving in the entire system.
The regression model is chosen based on the $C_{p}$-type statistic
described in \citet{efron04lars}.
If there is no bottleneck, then the response times of jobs that are
assigned to different queues should be independent, so the coefficient
of the second covariate---number of jobs in the entire system---should
be zero. So if that coefficient is nonzero, we predict a bottleneck.
Despite the naturalness of this approach, it might perform poorly
because the relationship between the covariates and the response time
is highly nonlinear.

We consider an alternative approach based on a queueing network model.
We use a simple procedure for selection among nested families of
queueing models. The procedure
relies on the fact that commonly used families of service time
distributions, such as exponential, gamma, and lognormal, include
distributions that come arbitrarily close to putting all their mass at
zero. The method is to
start with a queueing network that represents the expected performance
of the system, based on the developers' prior knowledge. Then add a
single hypothesized queue to the network, called the bottleneck queue,
that represents a hypothetical bottleneck in the system.
The Gibbs sampler now yields a sequence $m_{1}, m_{2}, \ldots, m_{N}$
of mean service times for the bottleneck queue. Choosing between the
base model and the augmented model can be thought of as testing whether
the mean service time of the bottleneck queue is zero. To do this, we
use the test statistic
$z = N^{-1} \sum_{i=1}^{N} m_{i}/\sigma$, where $\sigma$ is the standard
deviation of the $m_{i}$. This statistic is asymptotically  normal.
An alternative approach to model selection might rely on directly computing
the likelihood, but computing this quantity is notoriously difficult in
the queueing
setting, even for models that are much simpler than ours.

For the purposes of demonstrating the technique, we use a simple search
through model space, in which we hypothesize a bottleneck involving two
of the five queues, as in Figure~\ref{fig:alternatives}(b). Ten
possible alternative models are considered, each corresponding to a
different pair of work queues being connected to the bottleneck. For
each possible network, we test the hypothesis that the mean of the
bottleneck queue is zero, as described above. The result of the test is
counted as correct if the null is accepted and the true network is
Figure~\ref{fig:alternatives}(a), or if the null is rejected and the
true network is Figure~\ref{fig:alternatives}(c). The confidence level
used is 0.025.

We test both LARS and the queueing model-based technique on synthetic
data generated from the models in
Figure~\ref{fig:alternatives}(a) and~(b).
For both models, the arrival process is a homogeneous Poisson process
with parameter $\lambda= 1$. The service-time distribution of the work
queues is exponential with mean 2.5, so that each work queue has
utilization 0.5. For the model in Figure~\ref{fig:alternatives}(b),
the utilization of the bottleneck queue is varied in $\{ 0.001, 0.1,
0.25, 0.5, 0.75 \}$. Each synthetic data set contains 100 tasks. The
mean parameters of the service-time distributions are resampled in a
Bayesian fashion, using the prior described in Section~\ref{sec:network}.
The sampler is run for 5000 iterations. This experiment is repeated 5
times on independently generated synthetic data sets.

%
\begin{table}
\tabcolsep=0pt
\tablewidth=226pt
\caption{Error on missing-queue selection problem, as a function of
the utilization of the bottleneck queue. Lower utilization makes the
model selection problem harder. Chance performance is $0.5$}
\label{tbl:model-selection}
\begin{tabular*}{226pt}{@{\extracolsep{\fill}}ld{1.2}c@{}}
\hline
& \multicolumn{1}{c}{\textbf{Error}} & \textbf{Error} \\
\textbf{Utilization} & \multicolumn{1}{c}{\textbf{(queueing model)}} & \textbf{(linear model)} \\
\hline
0.001 & 0.50 & 0.55 \\
0.100 & 0.43 & 0.54 \\
0.250 & 0.28 & 0.49 \\
0.500 & 0.02 & 0.49 \\
0.750 & 0 & 0.43 \\
\hline
\end{tabular*}
\end{table}

\begin{figure}

\includegraphics{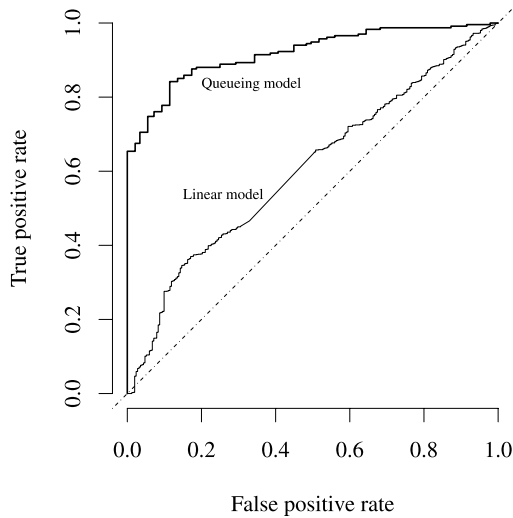}

\caption{Performance on missing-queue detection (ROC curve).}
\label{fig:missing-roc}
\end{figure}

The performance of the two techniques is shown in Table~\ref
{tbl:model-selection}. We report the percentage of correct missing
queue decisions, as a function of the utilization of the bottleneck
queue. When the utilization of the bottleneck queue is high, the
missing queue should be easy to detect. The LARS-based method performs
very poorly, performing only slightly better than chance (chance is 50\%) even on the easy cases. The queueing model technique, on the other
hand, performs perfectly on the easy cases, and does progressively
worse as the problem becomes harder. Figure~\ref{fig:missing-roc}
displays the same data as an ROC curve, generated using the R package
of \citet{rocr}. The model selection method has an area under the ROC
curve of 0.92, while that for the LARS-based method is 0.57.

\section{Discussion}
\label{sec:discussion}

In this paper we have introduced a novel perspective on queueing
networks that allows
inference in the presence of missing data.
The main idea is that a queueing model defines a deterministic transformation
between service times, which are independent, to the measured departure
times, which are highly dependent. This perspective allowed us to
develop an MCMC
sampler for the posterior distribution over the missing departure time data.
To our knowledge, this is the first example of inference in networks of
queues with missing data.
We demonstrated the effectiveness of this approach on data from an
actual Web application.

%

The fact that queueing networks are natural models of distributed
systems is
attested to by a large literature on these models in computer science.
For example, previous work has considered
queueing network models of single computer systems [\citet
{Lazowskaqsp}], computer networks [\citet{kleinrock}],
distributed file systems [\citet{ironmodel}], and Web applications
[\citet{urgaonkar05nalytical}; \citet{welshthesis}].
Our work builds on this literature, providing a statistical perspective
on networks of queues.

Within the statistical literature, inference in single-queue models has
been considered in both frequentist and Bayesian settings [\citet
{armero94mm1}; \citet{bhat97statistical}; \citet{insua98queueing}]. Previous work has
focused on single queues rather than networks, however [for
exceptions, see \citet{armero99queueing}, 
and   \citet{thir92queue}].
In addition, previous work has typically focused on a restricted class
of queueing disciplines and restricted patterns of missing data.
For example, one special case that has been considered is a
single-queue model in which all of the departure times are observed,
but none of the arrival times.
\citet{heggland2004} present an estimator for this problem based on
indirect inference.
\citet{FearnheadFiltering} presents a potentially more efficient
algorithm based on ideas similar to the embedded Markov chain from
queueing theory, while earlier \citet{jones99balking} presents a
similar algorithm that takes balking into account.
The indirect inference approach could possibly be extended to more
general situations, but it is difficult to see how to do so with the
dynamic programming approaches of Jones and
Fearnhead.
Another special setup that has been considered, again in single queues,
is the multi-step interdeparture distribution [\citet{luh99interdeparture}].

One issue that has been raised in the literature on Bayesian statistics
in queueing models is the effect of the prior. For example, as \citet
{Armero1994Prior} point out, in an exponential single-processor queue,
if gamma priors are placed on the service and arrival rates, then the
posterior moments of certain performance metrics, such as the
steady-state number of customers in a system, do not exist.
This is a disturbing issue if the goal of the analysis is to predict
the long-term behavior of the system. In diagnostic settings, however,
such as those in Section~\ref{sec:diagnosis}, this is a less serious
issue, because our interest lies in estimating the service times $\bs$
of individual jobs. The moments of the distribution $p(\bs| O, \theta
)$ do exist, even if the system is asymptotically unstable. That having
been said, it would not be difficult to modify our sampler to
incorporate more sophisticated priors on $\theta$ that address this
issue, such as those of \citet{Armero1994Prior}.

Another research area that is related to the current work is network
tomography [\citet{castro04tomography}; \citet{coates02tomography}], which
focuses on problems such as estimating the delays on each link of a
network solely from measurements of the end-to-end delay. This is a
markedly different inferential problem from ours, in that the network
tomography literature does not focus on how much of the link delay is
caused by the load on that link. For this reason, in our setting the
observed data always includes the number of requests in the system, a
measurement that is usually assumed to be unavailable in the network
tomography setup.

Finally, the current work suggests several largely unexplored
directions for future research. One direction concerns extensions to
the queueing models themselves, such as using a generalized linear
model in the service distribution. Another direction involves a
hierarchical prior on $\theta$, for example, to model the fact that
some queues may be known to run similar software and hardware, and
therefore should have similar performance characteristics. It would
also be interesting to examine the effects of the prior and of the
choice of service distribution on data imputation. Finally, at a higher
level, this work can be viewed as a coarse-grained generative model of
computer performance, and more detailed models could be of significant interest.
More information about the current state of the art in large-scale Web
applications can be found in a recent book by two leading engineers at
Google [\citet{barrosobook}].


\printaddresses


\begin{thebibliography}{99}
%
%
\bibitem[\protect\citeauthoryear{Armero and Bayarri}{1994a}]{armero94mm1}
\begin{barticle}[author]
\bauthor{\bsnm{Armero},~\bfnm{C.}\binits{C.}} \AND
\bauthor{\bsnm{Bayarri},~\bfnm{M.~J.}\binits{M.~J.}}
(\byear{1994}a).
\btitle{Bayesian prediction in {M$/$M$/$1} queues}.
\bjournal{Queueing Systems}
\bvolume{15}
\bpages{401--418}.
\end{barticle}
\MR{1266803}
\endbibitem

%
%
\bibitem[\protect\citeauthoryear{Armero and Bayarri}{1994b}]{Armero1994Prior}
\begin{barticle}[author]
\bauthor{\bsnm{Armero},~\bfnm{C.}\binits{C.}} \AND
\bauthor{\bsnm{Bayarri},~\bfnm{M.~J.}\binits{M.~J.}}
(\byear{1994}b).
\btitle{Prior assessments for prediction in queues}.
\bjournal{J. Roy. Statist. Soc. Ser.~D}
\bvolume{43}
\bpages{139--153}.
\bid{doi={10.2307/2348939}}
\end{barticle}
\endbibitem

%
%
\bibitem[\protect\citeauthoryear{Armero and Bayarri}{1999}]{armero99queueing}
\begin{bincollection}[author]
\bauthor{\bsnm{Armero},~\bfnm{C.}\binits{C.}} \AND
\bauthor{\bsnm{Bayarri},~\bfnm{M.~J.}\binits{M.~J.}}
(\byear{1999}).
\btitle{Dealing with uncertainties in queues and networks of queues: A
{Bayesian} approach}.
In \bbooktitle{Multivariate Analysis, Design of Experiments and Survey
Sampling}
(\beditor{\bfnm{S.}\binits{S.}~\bsnm{Ghosh}}, ed.)
\bpages{579--608}.
\bpublisher{Dekker}, \baddress{New York}.
\end{bincollection}
\MR{1722512}
\endbibitem

%
%
\bibitem[\protect\citeauthoryear{Barroso and H{\"o}lzle}{2009}]{barrosobook}
\begin{bbook}[author]
\bauthor{\bsnm{Barroso},~\bfnm{Luiz~Andr{\'e}}\binits{L.~A.}} \AND
\bauthor{\bsnm{H{\"o}lzle},~\bfnm{Urs}\binits{U.}}
(\byear{2009}).
\btitle{The Datacenter as a Computer: An Introduction to the Design of
Warehouse-Scale Machines}.
\bseries{Synthesis Lectures on Computer Architecture}.
\bpublisher{Morgan and Claypool},
San Rafael, CA.
\end{bbook}
\endbibitem

%
%
\bibitem[\protect\citeauthoryear{Bhat, Miller and
Rao}{1997}]{bhat97statistical}
\begin{bincollection}[author]
\bauthor{\bsnm{Bhat},~\bfnm{U.~Narayan}\binits{U.~N.}},
\bauthor{\bsnm{Miller},~\bfnm{Gregory~K.}\binits{G.~K.}} \AND
\bauthor{\bsnm{Rao},~\bfnm{S.~Subba}\binits{S.~S.}}
(\byear{1997}).
\btitle{Statistical analysis of queueing systems}.
In \bbooktitle{Frontiers in Queueing: Models and Applications in
Science and
Engineering}
\bpages{351--394}.
\bpublisher{CRC Press}, \baddress{Boca Raton, FL}.
\end{bincollection}
\MR{1440196}
\endbibitem

%
%
\bibitem[\protect\citeauthoryear{Castro
et al.}{2004}]{castro04tomography}
\begin{barticle}[author]
\bauthor{\bsnm{Castro},~\bfnm{R.}\binits{R.}},
\bauthor{\bsnm{Coates},~\bfnm{M.}\binits{M.}},
\bauthor{\bsnm{Liang},~\bfnm{G.}\binits{G.}},
\bauthor{\bsnm{Nowak},~\bfnm{R.}\binits{R.}} \AND
\bauthor{\bsnm{Yu},~\bfnm{B.}\binits{B.}}
(\byear{2004}).
\btitle{Network tomography: Recent developments}.
\bjournal{Statist. Sci.}
\bvolume{19}
\bpages{499--517}.
\end{barticle}
\MR{2185628}
\endbibitem

%
%
\bibitem[\protect\citeauthoryear{Celeux and Diebolt}{1985}]{celeux85sem}
\begin{barticle}[author]
\bauthor{\bsnm{Celeux},~\bfnm{Gilles}\binits{G.}} \AND
\bauthor{\bsnm{Diebolt},~\bfnm{Jean}\binits{J.}}
(\byear{1985}).
\btitle{The {SEM} algorithm: {A} probabilistic teacher algorithm
derived from
the {EM} algorithm for the mixture problem}.
\bjournal{Comput. Statist. Quart.}
\bvolume{2}
\bpages{73--82}.
\end{barticle}
\endbibitem

%
%
\bibitem[\protect\citeauthoryear{Coates et al.}{2002}]{coates02tomography}
\begin{barticle}[author]
\bauthor{\bsnm{Coates},~\bfnm{M.}\binits{M.}},
\bauthor{\bsnm{Hero},~\bfnm{A.}\binits{A.}},
\bauthor{\bsnm{Nowak},~\bfnm{R.}\binits{R.}} \AND
\bauthor{\bsnm{Yu},~\bfnm{B.}\binits{B.}}
(\byear{2002}).
\btitle{Internet tomography}.
\bjournal{IEEE Signal Processing Magazine}
\bvolume{19}
\bpages{47--65}.
\end{barticle}
\endbibitem

%
%
\bibitem[\protect\citeauthoryear{Cormen et al.}{2001}]{clrs}
\begin{bbook}[author]
\bauthor{\bsnm{Cormen},~\bfnm{T.~H.}\binits{T.~H.}},
\bauthor{\bsnm{Leiserson},~\bfnm{C.~E.}\binits{C.~E.}},
\bauthor{\bsnm{Rivest},~\bfnm{R.~L.}\binits{R.~L.}} \AND
\bauthor{\bsnm{Stein},~\bfnm{C.}\binits{C.}}
(\byear{2001}).
\btitle{Introduction to Algorithms}.
\bpublisher{MIT Press}, \baddress{Cambridge, MA}.
\end{bbook}
\MR{1848805}
\endbibitem

%
%
\bibitem[\protect\citeauthoryear{Damien, Wakefield and
Walker}{1999}]{damien99gibbs}
\begin{barticle}[author]
\bauthor{\bsnm{Damien},~\bfnm{P.}\binits{P.}},
\bauthor{\bsnm{Wakefield},~\bfnm{J.~C.}\binits{J.~C.}} \AND
\bauthor{\bsnm{Walker},~\bfnm{S.~G.}\binits{S.~G.}}
(\byear{1999}).
\btitle{Gibbs sampling for {Bayesian} nonconjugate and hierarchical
models by
using auxiliary variables}.
\bjournal{J. R. Stat. Soc. Ser. B Stat. Methodol.}
\bvolume{61}
\bpages{331--344}.
\end{barticle}
\MR{1680334}
\endbibitem

%
%
\bibitem[\protect\citeauthoryear{Efron et al.}{2004}]{efron04lars}
\begin{barticle}[author]
\bauthor{\bsnm{Efron},~\bfnm{Bradley}\binits{B.}},
\bauthor{\bsnm{Hastie},~\bfnm{Trevor}\binits{T.}},
\bauthor{\bsnm{Johnstone},~\bfnm{Iain}\binits{I.}} \AND
\bauthor{\bsnm{Tibshirani},~\bfnm{Robert}\binits{R.}}
(\byear{2004}).
\btitle{Least angle regression}.
\bjournal{Ann. Statist.}
\bvolume{32}
\bpages{407--499}.
\end{barticle}
\MR{2060166}
\endbibitem

%
%
\bibitem[\protect\citeauthoryear{Fearnhead}{2004}]{FearnheadFiltering}
\begin{barticle}[author]
\bauthor{\bsnm{Fearnhead},~\bfnm{Paul}\binits{P.}}
(\byear{2004}).
\btitle{Filtering recursions for calculating likelihoods for queues
based on
inter-departure time data}.
\bjournal{Statist. Comput.}
\bvolume{14}
\bpages{261--266}.
\end{barticle}
\MR{2086402}
\endbibitem

%
%
\bibitem[\protect\citeauthoryear{Heggland and Frigessi}{2004}]{heggland2004}
\begin{barticle}[author]
\bauthor{\bsnm{Heggland},~\bfnm{Knut}\binits{K.}} \AND
\bauthor{\bsnm{Frigessi},~\bfnm{Arnoldo}\binits{A.}}
(\byear{2004}).
\btitle{Estimating functions in indirect inference}.
\bjournal{J. R. Stat. Soc. Ser. B Stat. Methodol.}
\bvolume{66}
\bpages{447--462}.
\end{barticle}
\MR{2062387}
\endbibitem

%
%
\bibitem[\protect\citeauthoryear{Insua, Wiper and
Ruggeri}{1998}]{insua98queueing}
\begin{barticle}[author]
\bauthor{\bsnm{Insua},~\bfnm{David~R.}\binits{D.~R.}},
\bauthor{\bsnm{Wiper},~\bfnm{Michael}\binits{M.}} \AND
\bauthor{\bsnm{Ruggeri},~\bfnm{Fabrizio}\binits{F.}}
(\byear{1998}).
\btitle{Bayesian analysis of {M$/$Er$/$1} and {M$/$H\_k$/$1} queues}.
\bjournal{Queueing Syst. Theory Appl.}
\bvolume{30}
\bpages{289--308}.
\bid{doi={10.1023/A:1019173206509}}
\end{barticle}
\MR{1672243}
\endbibitem

%
%
\bibitem[\protect\citeauthoryear{Jones}{1999}]{jones99balking}
\begin{barticle}[author]
\bauthor{\bsnm{Jones},~\bfnm{Lee~K.}\binits{L.~K.}}
(\byear{1999}).
\btitle{Inferring balking behavior from transactional data}.
\bjournal{Oper. Res.}
\bvolume{47}
\bpages{778--784}.
\end{barticle}
\endbibitem

%
%
\bibitem[\protect\citeauthoryear{Kleinrock}{1973}]{kleinrock}
\begin{bbook}[author]
\bauthor{\bsnm{Kleinrock},~\bfnm{L.}\binits{L.}}
(\byear{1973}).
\btitle{Queueing Systems: Theory and Applications}.
\bpublisher{Wiley Interscience}, \baddress{New York}.
\end{bbook}
\endbibitem

%
%
\bibitem[\protect\citeauthoryear{Lauritzen}{1996}]{lauritzenbook}
\begin{bbook}[author]
\bauthor{\bsnm{Lauritzen},~\bfnm{Steffen~L.}\binits{S.~L.}}
(\byear{1996}).
\btitle{Graphical Models}.
\bpublisher{{Oxford Univ. Press}}.
\end{bbook}
\MR{1419991}
\endbibitem

%
%
\bibitem[\protect\citeauthoryear{Lazowska et al.}{1984}]{Lazowskaqsp}
\begin{bbook}[author]
\bauthor{\bsnm{Lazowska},~\bfnm{Edward~D.}\binits{E.~D.}},
\bauthor{\bsnm{Zahorjan},~\bfnm{John}\binits{J.}},
\bauthor{\bsnm{Graham},~\bfnm{G.~Scott}\binits{G.~S.}} \AND
\bauthor{\bsnm{Sevcik},~\bfnm{Kenneth~C.}\binits{K.~C.}}
(\byear{1984}).
\btitle{Quantitative System Performance: Computer System Analysis Using
Queueing Network Models}.
\bpublisher{Prentice Hall, Englewood Cliffs, NJ.}
\end{bbook}
\endbibitem

%
%
\bibitem[\protect\citeauthoryear{Luh}{1999}]{luh99interdeparture}
\begin{barticle}[author]
\bauthor{\bsnm{Luh},~\bfnm{H.}\binits{H.}}
(\byear{1999}).
\btitle{Derivation of the $N$-step interdeparture time distribution in {GI$/$G$/$1}
queueing systems}.
\bjournal{Eur. J. Oper. Res.}
\bvolume{118}
\bpages{194--212}.
\end{barticle}
\endbibitem

%
%
\bibitem[\protect\citeauthoryear{Neal}{2003}]{Neal2003Slice}
\begin{barticle}[author]
\bauthor{\bsnm{Neal},~\bfnm{Radford~M.}\binits{R.~M.}}
(\byear{2003}).
\btitle{Slice sampling}.
\bjournal{Ann. Statist.}
\bvolume{31}
\bpages{705--741}.
\end{barticle}
\MR{1994729}
\endbibitem

%
%
\bibitem[\protect\citeauthoryear{Sing et al.}{2005}]{rocr}
\begin{barticle}[author]
\bauthor{\bsnm{Sing},~\bfnm{Tobias}\binits{T.}},
\bauthor{\bsnm{Sander},~\bfnm{Oliver}\binits{O.}},
\bauthor{\bsnm{Beerenwinkel},~\bfnm{Niko}\binits{N.}} \AND
\bauthor{\bsnm{Lengauer},~\bfnm{Thomas}\binits{T.}}
(\byear{2005}).
\btitle{{ROCR:} Visualizing classifier performance in {R}}.
\bjournal{Bioinformatics}
\bvolume{21}
\bpages{3940--3941}.
\end{barticle}
\endbibitem

%
%
\bibitem[\protect\citeauthoryear{Sobel et al.}{2008}]{Cloudstone}
\begin{binproceedings}[author]
\bauthor{\bsnm{Sobel},~\bfnm{Will}\binits{W.}},
\bauthor{\bsnm{Subramanyam},~\bfnm{Shanti}\binits{S.}},
\bauthor{\bsnm{Sucharitakul},~\bfnm{Akara}\binits{A.}},
\bauthor{\bsnm{Nguyen},~\bfnm{Jimmy}\binits{J.}},
\bauthor{\bsnm{Wong},~\bfnm{Hubert}\binits{H.}},
\bauthor{\bsnm{Patil},~\bfnm{Sheetal}\binits{S.}},
\bauthor{\bsnm{Fox},~\bfnm{Armando}\binits{A.}} \AND
\bauthor{\bsnm{Patterson},~\bfnm{David}\binits{D.}}
(\byear{2008}).
\btitle{{Cloudstone}: Multi-platform, multi-language benchmark and measurement
tools for {Web} 2.0}.
In \bbooktitle{First Workshop on Cloud Computing and its Applications ({CCA})},
Chicago, IL.
\end{binproceedings}
\endbibitem

%
%
\bibitem[\protect\citeauthoryear{Thereska and Ganger}{2008}]{ironmodel}
\begin{binproceedings}[author]
\bauthor{\bsnm{Thereska},~\bfnm{Eno}\binits{E.}} \AND
\bauthor{\bsnm{Ganger},~\bfnm{Gregory~R.}\binits{G.~R.}}
(\byear{2008}).
\btitle{IRONModel: Robust performance models in the wild}.
In \bbooktitle{SIGMETRICS}
\bvolume{36}
\bpages{253--264}.
\bpublisher{ACM},
\baddress{New York}.
\end{binproceedings}
\endbibitem

%
%
\bibitem[\protect\citeauthoryear{Thiruvaiyaru and Basawa}{1992}]{thir92queue}
\begin{barticle}[author]
\bauthor{\bsnm{Thiruvaiyaru},~\bfnm{Dharma}\binits{D.}} \AND
\bauthor{\bsnm{Basawa},~\bfnm{Ishwar~V.}\binits{I.~V.}}
(\byear{1992}).
\btitle{Empirical Bayes estimation for queueing systems and networks}.
\bjournal{Queueing Systems}
\bvolume{11}
\bpages{179--202}.
\bid{doi={10.1007/BF01163999}}
\end{barticle}
\MR{1184091}
\endbibitem

%
%
\bibitem[\protect\citeauthoryear{Urgaonkar et al.}{2005}]{urgaonkar05nalytical}
\begin{binproceedings}[author]
\bauthor{\bsnm{Urgaonkar},~\bfnm{Bhuvan}\binits{B.}},
\bauthor{\bsnm{Pacifici},~\bfnm{Giovanni}\binits{G.}},
\bauthor{\bsnm{Shenoy},~\bfnm{Prashant}\binits{P.}},
\bauthor{\bsnm{Spreitzer},~\bfnm{Mike}\binits{M.}} \AND
\bauthor{\bsnm{Tantawi},~\bfnm{Assar}\binits{A.}}
(\byear{2005}).
\btitle{An analytical model for multi-tier {Internet} services and its
applications}.
In \bbooktitle{SIGMETRICS}
\bvolume{33}
\bpages{291--302}.
\bpublisher{ACM},
\baddress{New York}.
\end{binproceedings}
\endbibitem

%
%
\bibitem[\protect\citeauthoryear{Welsh}{2002}]{welshthesis}
\begin{bphdthesis}[author]
\bauthor{\bsnm{Welsh},~\bfnm{Matt}\binits{M.}}
(\byear{2002}).
\btitle{An Architecture for Highly Concurrent, Well-Conditioned Internet
Services}
\btype{Ph.D. thesis}, \bschool{Univ. California, Berkeley}.
\end{bphdthesis}
\endbibitem
\end{thebibliography}
\end{document}